\documentclass[journal]{IEEEtran}

\usepackage{lineno,hyperref}
\usepackage{bm}
\usepackage{mathrsfs}
\usepackage{amsmath}
\usepackage{amssymb}
\usepackage{graphicx}
\usepackage{subfigure}
\usepackage{epstopdf}
\usepackage{cite}
\usepackage{xcolor}
\usepackage{wasysym}
\usepackage{footnote}
\usepackage{amssymb}
\usepackage{overpic}
\modulolinenumbers[5]

\ifCLASSINFOpdf

\else

\fi

\begin{document}

\title{Robust DCD-Based Recursive Adaptive Algorithms}

\author{Yi~Yu,
        ~Lu Lu,~\IEEEmembership{Member,~IEEE},
        ~Zongsheng Zheng,~\IEEEmembership{Student Member,~IEEE},
        ~Wenyuan Wang,\\
        ~Yuriy Zakharov,~\IEEEmembership{Senior Member,~IEEE},
        ~and
        ~Rodrigo C. de Lamare,~\IEEEmembership{Senior Member,~IEEE}
\thanks{This work was supported by National Science Foundation of P.R. China (Nos. 61901400 and 61901285) and Doctoral Research Fund of Southwest University of Science and Technology in China (No. 19zx7122). The work of Y. Zakharov was partly supported by the U.K. EPSRC Grants EP/P017975/1 and EP/R003297/1.}

\thanks{Y. Yu is with School of Information Engineering, Robot Technology Used for Special Environment Key Laboratory of Sichuan Province, Southwest University of Science and Technology, Mianyang, 621010, China (e-mail: yuyi\_xyuan@163.com).}

\thanks{L. Lu is with School of Electronics and Information Engineering, Sichuan University, Chengdu, China (e-mail: lulu19900303@126.com).}

\thanks{Z. Zheng and W. Wang are with School of Electrical Engineering, Southwest Jiaotong University, Chengdu, 610031, China (e-mail: wenyuanwang@my.swjtu.edu.cn, bk20095185@my.swjtu.edu.cn).}

\thanks{Y. Zakharov is with the Department of Electronics, University of York, York YO10 5DD, U.K. (e-mail: yury.zakharov@york.ac.uk).}

\thanks{R. C. de Lamare is with CETUC, PUC-Rio, Rio de Janeiro 22451-900, Brazil. (e-mail: delamare@cetuc.puc-rio.br).}
}


\maketitle

\begin{abstract}
The dichotomous coordinate descent (DCD) algorithm has been successfully used for significant reduction in the complexity of recursive least squares (RLS) algorithms. In this work, we generalize the application of the DCD algorithm to RLS adaptive filtering in impulsive noise scenarios and derive a unified update formula. By employing different robust strategies against impulsive noise, we develop novel computationally efficient DCD-based robust recursive algorithms. Furthermore, to equip the proposed algorithms with the ability to track abrupt changes in unknown systems, a simple variable forgetting factor mechanism is also developed. Simulation results for channel identification scenarios in impulsive noise demonstrate the effectiveness of the proposed algorithms.
\end{abstract}


\IEEEpeerreviewmaketitle

\section{Introduction}

\IEEEPARstart{a}{daptive} filtering has been a prominent technique
in a variety of applications such as system identification, active
noise control, and echo cancellation (EC)
\cite{sayed2003fundamentals}. The least mean square (LMS) and
recursive least squares (RLS) algorithms represent two typical
families of adaptive algorithms
\cite{ifir},\cite{jio},\cite{smtvb},\cite{jidf},\cite{jiols},\cite{saalt}.
The complexity of LMS is $\mathcal{O}(M)$ arithmetic operations per
sample (ops), where $M$ is the filter length, but its convergence is
slow especially when the input signal is highly correlated. RLS
improves the convergence at the cost of a high complexity of
$\mathcal{O}(M^2)$ ops. To reduce the complexity, some fast RLS
algorithms were proposed as summarized in
\cite[Chapter~14]{sayed2003fundamentals}. However, these fast
algorithms are numerically unstable in finite precision
implementation since they are based on the matrix inversion.

Alternatively, the dichotomous coordinate descent (DCD) iterations
for solving the normal equations in the RLS algorithms were proposed
\cite{zakharov2008RLS}. They result in not only numerically stable
adaptive algorithms but also in performance comparable to that of
the original RLS algorithm. An important property of the DCD
algorithm is that it only requires addition and shift operations,
which are simpler for implementation than multiplication and
division, and thus it is well suited to real-time implementation.
Moreover, the DCD-RLS algorithm reduces the complexity to
$\mathcal{O}(M)$ ops for input signals with the tapped-delay
structure. The DCD algorithm was also applied for the complexity
reduction in the affine projection algorithm~\cite{zakharov2008low},
sparse signal recovery~\cite{zakharov2017low}, and distributed
estimation~\cite{yu2019robust}.

Regrettably, the LMS and RLS algorithms undergo performance
deterioration in impulsive noise~\cite{ZHANG201467}, owing to the
squared-error based minimization criteria. Realizations of impulsive
noise process are sparse and random with amplitude far higher than
the Gaussian noise, and therefore, best modeled by heavy-tailed
distributions, e.g., the $\alpha$-stable distribution. Such noise
scenarios are common in such as echo cancellation, underwater
acoustics, audio processing, array processing, distributed
processing and
communications~\cite{nikias1995signal},\cite{zimmermann2002analysis},\cite{sjidf},\cite{jidf_echo},\cite{jioccm},\cite{dce},\cite{als},\cite{damdc}.
For adapting impulsive noise scenarios, existing literature have
reported various robust approaches \cite{rccm},\cite{locsme,okspme}.
For instance, the recursive least M-estimate (RLM)
algorithm~\cite{zou2001robust} exploits the Hampel's M-estimate
function to suppress impulsive interferences. Based on the
$l_p$-norm of errors, the recursive least $p$-norm (RL$p$N)
algorithm was developed~\cite{navia2012combination}. By gathering
all the $p$-norms from $p=1$ to 2 of the error, the continuous mixed
$p$-norms~(CMPN) algorithm was derived~\cite{zayyani2014continuous};
however, it has slow convergence for correlated inputs due to the
gradient descent (GD) principle. Taking advantage of the
Geman-McClure (GMC) estimator, a recursive
algorithm~\cite{lu2018recursive} for Volterra system identification
was derived, which shows a better performance than RL$p$N and RLM
algorithms in impulsive noise modeled by the $\alpha$-stable
distribution \cite{nikias1995signal}. When impulsive noise appears,
by incorporating the step-size scaler into the update term, a robust
subband algorithm was developed~\cite{hur2016variable}. The
correntropy measures the similarity between two variables, which is
helpful for suppressing large outliers; thus, the maximum
correntropy criterion (MCC) has been used for improving the
anti-jamming capability of adaptive filters to impulsive noise,
yielding the GD-based
MCC~\cite{chen2016generalized,shi2018improved,chen2014steady} and
recursive MCC (RMCC)
algorithms~\cite{chen2015convergence,radmanesh2018recursive}.
However, these robust recursive algorithms have also high complexity
of $\mathcal{O}(M^2)$ ops. In particular, the complexity of the
fixed-point variant of MCC algorithm in~\cite{chen2015convergence}
is $\mathcal{O}(M^3)$ due to the direct inverse of an $M\times M$
matrix.

This work focuses on a class of low-complexity robust algorithms
against impulsive noise by resorting to the DCD approach.
Concretely, a generalized DCD-based robust recursion is derived. By
applying different robust strategies to this recursion, we develop
DCD-based robust algorithms, such as the DCD-RMCC, DCD-RLM, and
DCD-RL$p$N algorithms. We also design a variable forgetting factor
(VFF) scheme for improving the tracking capability of the
algorithms.

\section{DCD-Based Robust Algorithms}
\subsection{Unified Formulation}
Suppose that at time instant $n$, the desired signal $d_n$ and an $M\times1$ input signal vector $\bm x_n$ are available and obey the relation $d_n = \bm x_n^\text{T} \bm w^o+v_n$, where the $M\times1$ vector $\bm w^o$ needs to be estimated, and $(\cdot)^\text{T}$ denotes the transpose. The additive noise with impulsive behavior, $v_n$, here is described by the $\alpha$-stable process\footnote{Other models describing the noise with impulses include the contaminated-Gaussian (CG) model \cite{yu2019robust} and the Gaussian mixture model (GMM) \cite{kassam2012signal}.}, also called the $\alpha$-stable noise. A (symmetric) $\alpha$-stable random variable is usually characterized by the characteristic function~\cite{nikias1995signal}
\begin{align}
\label{001}
\phi(t)=\exp(-\gamma \lvert t \lvert^\alpha).
\end{align}
The characteristic exponent $\alpha \in (0,2]$ describes the impulsiveness of the noise (smaller $\alpha$ leads to more outliers) and $\gamma>0$ represents the dispersion degree of the noise. Note that when $\alpha$ = 1 and~2, it reduces to the Cauchy and Gaussian distributions, respectively.

To effectively estimate $\bm w^o$ in such noise scenarios, we define a unified robust exponentially weighted least squares problem:
\begin{align}
\bm w_n = \arg \min\limits_{\bm w} \left\lbrace \sum\limits_{i=0}^n \lambda^{n-i} \varphi \left(d_i-\bm x_{i}^\text{T}\bm w\right) + \delta_n \lVert \bm w\rVert_2^2 \right\rbrace,
\label{002}
\end{align}
where $0\ll \lambda<1$ is the forgetting factor, $\delta_n>0$ is a regularization parameter, and $\varphi(\cdot)$ is a function that specifies the robustness against impulsive noise.

By setting the derivative of \eqref{002} with respect to $\bm w$ to zero, we arrive at the normal equations:
\begin{align}
\bm R_n \bm w_n = \bm z_n,
\label{003}
\end{align}
where
\begin{equation}
\begin{array}{rcl}
\begin{aligned}
\bm R_n & = \sum\limits_{i=0}^n \lambda^{n-i} f_i \bm x_i \bm x_i^\text{T} + \delta_n \bm I_M\\
&= \lambda \bm R_{n-1}+f_n \bm x_n \bm x_n^\text{T} + (\delta_n - \lambda \delta_{n-1})\bm I_M
\label{004}
\end{aligned}
\end{array}
\end{equation}
is the time-averaged autocorrelation matrix of $\bm x_n$,
\begin{equation}
\begin{array}{rcl}
\begin{aligned}
\bm z_n &= \sum\limits_{i=0}^n \lambda^{n-i} f_i d_i \bm x_i \\
&= \lambda \bm z_{n-1}+f_n d_n \bm x_n
\label{005}
\end{aligned}
\end{array}
\end{equation}
is the time-averaged crosscorrelation vector of $d_n$ and $\bm x_n$, and $\bm I_M$ is an $M \times M$ identity matrix. Also, $f_n=\varphi'(\epsilon_n)/\epsilon_n$, where $\epsilon_n=d_n-\bm x_{n}^\text{T}\bm w_n$ is the \emph{a posteriori} error and $\varphi'(\epsilon_n)$ is the derivative of $\varphi(\epsilon_n)$.

At time instant $n-1$, let $\hat{\bm w}_{n-1}$ denote the approximate solution of \eqref{003} for estimating $\bm w^o$, and the corresponding residual vector is $\bm r_{n-1} = \bm z_{n-1} - \bm R_{n-1} \hat{\bm w}_{n-1}$. By defining $\Delta \bm w_n = \bm w_n - \hat{\bm w}_{n-1}$, from \eqref{003} we obtain an auxiliary system of equations:
\begin{equation}
\begin{array}{rcl}
\begin{aligned}
\bm R_n \Delta \bm w_n = \bm z_n - \bm R_n \hat{\bm w}_{n-1} \triangleq \bm b_n.
\label{006}
\end{aligned}
\end{array}
\end{equation}
Applying the recursive expressions \eqref{004} and \eqref{005}, $\bm b_n$ can be rewritten as
\begin{align}
\bm b_n = \lambda \bm r_{n-1}+f_n e_n \bm x_n -(\delta_n - \lambda \delta_{n-1})\hat{\bm w}_{n-1},
\label{007}
\end{align}
where $e_n = d_n-\bm x_{n}^\text{T} \hat{\bm w}_{n-1}$ denotes the \emph{a priori} error.

By using the DCD algorithm to solve the problem in~\eqref{006}, we arrive at an approximate solution of the original normal equations~\eqref{003}:
\begin{align}
\hat{\bm w}_{n} = \hat{\bm w}_{n-1} + \Delta \hat{\bm w}_n.
\label{008}
\end{align}

Although \eqref{007} shows that $\bm b_n$ requires the residual error vector of the original system~\eqref{003}, after some algebra we notice that it is equivalent to the residual error vector for the auxiliary system~\eqref{006}, i.e., $\bm r_{n} = \bm z_{n} - \bm R_{n} \hat{\bm w}_{n} = \bm b_n - \bm R_n \Delta \hat{\bm w}_n$. At time index~$n$, $f_n$ in~\eqref{007} is not yet available, but by resorting to the \emph{a priori} error, we may approximate $f_n$ as
 \begin{align}
 f_n \approx \varphi'(e_n)/e_n.
 \label{009}
 \end{align}
\begin{table}[tbp]
    \scriptsize
    \centering
    \caption{DCD-Based Robust Recursive Update}
    \label{table_1}
    \begin{tabular}{lc}
        \hline
        \text{Parameters:} $0\ll\lambda<1,\;\delta_0>0$\\
        \text{Initialization:} $\bm R_0 = \delta_0 \bm I_M,\;\hat{\bm w}_0 = \bm 0,\;\bm r_0= \bm 0$\\
        \hline
        \text{for} $n=1,...$\\
        \;\;\;$e_n = d_n-\bm x_{n}^\text{T} \hat{\bm w}_{n-1}$\\
        \;\;\;$\bm R_n = \lambda \bm R_{n-1}+f_n \bm x_n \bm x_n^T + (\delta_n - \lambda \delta_{n-1})\bm I_M$\\
        \;\;\;$\bm b_n = \lambda \bm r_{n-1}+f_n e_n \bm x_n -(\delta_n - \lambda \delta_{n-1})\hat{\bm w}_{n-1}$\\
        \;\;\;\text{Using DCD iterations to solve} $\bm R_n \Delta \bm w_n = \bm b_n$, which yields $\Delta \hat{\bm w}_n$ and $\bm r_n$ \\
        \;\;\;$\hat{\bm w}_{n} = \hat{\bm w}_{n-1} + \Delta \hat{\bm w}_n$\\
        \text{end}\\
        \hline
    \end{tabular}
\end{table}
This completes the derivation of DCD-based robust recursion, summarized in Table~\ref{table_1}.

Table~\ref{table_2} presents the leading DCD algorithm for solving the system of equations $\bm R_n \Delta \bm w_n = \bm b_n$ (readers can refer to~\cite{zakharov2008low,zakharov2008RLS} for details), where $[\bm r_n]_l$ is the $l$-th entry of $\bm r_n$, and $[\bm R_n]_{l,l}$ and $[\bm R_n]_{:,l}$ are the $(l,l)$-th entry and the $l$-th column of $\bm R_n$, respectively. Herein, $[-H, H]$ denotes the amplitude range for elements of the solution vector $\Delta \hat{\bm w}_n$. It is often chosen as a power-of-two number so that all multiplications by $\mu$ can be implemented by bit-shifts. $M_b$ is the number of bits for a fixed-point representation of $\hat{\bm w}_n$ within the range $[-H, H]$. $N_u$ stands for a maximum number of elements in $\Delta \hat{\bm w}_n$ that are updated. The solution $\Delta \hat{\bm w}_n$ approaches the optimal one (i.e., $\Delta \hat{\bm w}_n=\bm R_n^{-1}\bm b_n$ ) as $N_u$ increases. As seen in Table~\ref{table_2}, the implementation of DCD only requires shift and addition operations, excluding multiplication and division operations.
\begin{table}[tbp]
    \scriptsize
    \centering
    \caption{Leading DCD Algorithm}
    \label{table_2}
    \begin{tabular}{lc}
        \hline
        \text{Parameters:} $H,\;N_u,\;M_b$,\\
        \text{Initialization:} $\Delta \hat{\bm w}_n = \bm 0,\;\bm r_n= \bm b_n,\;y=1,\;\mu=H/2$\\
        \hline
        \text{for} $j=1,...,N_u$\\
        \;\;\;$l= \arg \max \limits_{j=1,...,M} \{ \rvert [\bm r_n]_j \rvert \}$\\
        \;\;\;\text{while} $\rvert [\bm r_n]_l \rvert \leq (\mu/2) [\bm R_n]_{l,l}$ \text{and} $y \leq M_b$\\
        \;\;\;\;\;$y=y+1$,\;$\mu=\mu/2$ \\
        \;\;\;\text{end}\\
        \;\;\;\text{if} $y > M_b$\\
        \;\;\;\;\; \text{break}\\
        \;\;\;\text{else}\\
        \;\;\;\;\; $[\Delta \hat{\bm w}_n]_l \leftarrow [\Delta \hat{\bm w}_n]_l +\mu \text{sign}([\bm r_n]_l)$ \\
        \;\;\;\;\; $\bm r_n \leftarrow \bm r_n -\mu \text{sign}([\bm r_n]_l) [\bm R_n]_{:,l}$  \\
        \;\;\;\text{end}\\
        \text{end}\\
        \hline
    \end{tabular}
\end{table}

\subsection{Robust Strategies}
\begin{table}[tbp]
    \scriptsize
    \centering
    \caption{Some Robust DCD-based Algorithms}
    \label{table_3}
    \begin{tabular}{@{}lccc}
        \hline
        \text{Robust Algorithms} &$\varphi(e)$ in \eqref{002} &$f(e)=\varphi'(e)/e$ in \eqref{009}\\
        \hline
        DCD-RMCC &$\frac{1}{\sqrt{2\pi}\beta} \left[1- \exp \left( -\frac{e^2}{2\beta^2}\right)\right]$ &$\exp \left( -\frac{e^2}{2\beta^2}\right) $\\
        DCD-RLM & $ \left\{ \begin{aligned}
        &e^2/2, \text{ if } |e| \leq \xi\\
        &\xi^2/2, |e| > \xi,
        \end{aligned} \right.$ &$ \left\{ \begin{aligned}
        &1, \text{ if } |e| \leq \xi\\
        &0, |e| > \xi,
        \end{aligned} \right.$ \\
        DCD-RL$p$N &$|e|^p$ &$|e|^p/(|e|^2 + \varepsilon)$\\
        \hline
    \end{tabular}\\
\end{table}
Applying a particular robust strategy to define $\varphi(e)$ in~\eqref{002}, we can compute $f_n$ by \eqref{009} to arrive at a DCD-based robust algorithm. Table~\ref{table_3} gives examples of $\varphi(e)$ for the DCD-RMCC, DCD-RLM, and DCD-RL$p$N algorithms derived from the widely studied MCC, M-estimate, and $l_p$-norm strategies, respectively. We note the following about the proposed algorithms:

\textit{1)} For the DCD-RMCC algorithm, $\beta > 0$ denotes the kernel width. When $\beta\rightarrow \infty$, $f_n$ approaches 1 so that the DCD-RMCC algorithm reduces to the DCD-RLS algorithm. When $\beta\rightarrow 0$, $f_n$ becomes 0, and the DCD-RMCC update is frozen. Thus, $\beta$ balances the robustness and dynamic performance of the algorithm in impulsive noise.

\textit{2)} The DCD-RLM algorithm uses the modified Huber M-estimate function~\cite{chan2004recursive} for $\varphi(e)$\footnote{Other M-estimate functions may also be used, e.g., the Huber~\cite{petrus1999robust} and Hampel~\cite{zou2001robust} functions.}. When $|e_n| < \xi$, thus $f_n$ equals~1 so that the DCD-RLM algorithm becomes the DCD-RLS algorithm. Otherwise, $f_n$ becomes 0 to stop the update (ideally, this only happens when the impulsive noise appears). To effectively suppress the impulsive noise, the threshold $\xi$ is adaptively adjusted by $\xi =\tau \hat{\sigma}_{e,n}$,
\begin{align}
\label{010}
\hat{\sigma}_{e,n}^2 = \zeta \hat{\sigma}_{e,n-1}^2 + c_\sigma (1-\zeta) \text{med}(\bm a_n^e),
\end{align}
where $\left. 0 < \zeta <1\right. $ is a weighting factor (except $\zeta=0$ at the algorithm start), $\text{med}(\cdot)$ is the median operator which helps to remove outliers in the data window $\bm a_n^e=[e_n^2,e_{n-1}^2,...,e_{n-N_w+1}^2]$, and $c_\sigma = 1.483(1+5/(N_w-1))$ is the correction factor~\cite{zou2001robust}. It is worth noting that, the window length~$N_w$ should be properly chosen. Larger $N_w$ makes a more robust estimate $\hat{\sigma}_{e,n}^2$ from~\eqref{010} but requires a higher complexity. A typical value of $\tau$ is 2.576. If $e_n$ is assumed to be Gaussian (which is reasonable except when being polluted by impulsive noise), this value means the 99\% confidence to prevent $e_n$ from contributing to the update for $|e_n| \geq \xi$~\cite{zou2001robust}.

\textit{3)} The convergence of the RL$p$N algorithm in the $\alpha$-stable noise requires $0<p<\alpha$. If $p=2$, the DCD-RL$p$N algorithm will also become the DCD-RLS algorithm. When $p=1$, this corresponds to the recursion sign algorithm \cite{mathews1987improved} with good robustness against impulsive noise.

\emph{Remark 1:} In a nutshell, when impulsive noise happens, its negative influence on the updates of $\bm R_n$ and $\bm b_n$ will be lowered significantly due to by multiplying a tiny scaler~$f_n$ into the updates. Then, we can generalize the DCD recursion to find $\Delta \bm \hat{w}_n$ from the system of equations $\bm R_n \Delta \bm w_n = \bm b_n$ with impulse-free. Hence, according to~\eqref{008}, the proposed DCD-based algorithms can work well in impulsive noise.
\subsection{Computational Complexity}
The direct solution of~\eqref{003} is $\bm w_n = \bm R_n^{-1}\bm z_n$. The regularization $\delta_n$ is to maintain the numerical stability of this solution~\cite{sayed2003fundamentals}. However, this leads to the complexity of $\mathcal{O}(M^3)$ due to the matrix inversion $\bm R_n^{-1}$. Generally, $\delta_n$ is chosen as $\delta_n=\lambda^{n+1}\delta_0$ (e.g., in this paper), it makes \eqref{004} become $\bm R_n = \lambda \bm R_{n-1}+f_n \bm x_n \bm x_n^\text{T}$. Then, using the matrix inversion lemma, $\bm R_n^{-1}$ can be calculated in a recursive way so that the complexity of the resulting algorithm is $\mathcal{O}(M^2)$, while it is still high for large $M$.
\begin{table}[tbp]
    \scriptsize
    \centering
    \caption{Complexity of Algorithms per Input Sample}
    \label{table_4}
    \begin{tabular}{@{}l|c|c|cc}
        \hline
        \text{Algorithms} &\text{Additions} &\text{Multiplications} &\text{Divisions}\\
        \hline
        \text {LMS}  &$2M$ &$2M+1$ &0\\ \hline
        \text {(R) RLS}  &$3M^2+M$ &$4M^2+4M+1$ &1\\ \hline
        \begin{tabular}[c]{@{}l@{}} \text {(R) DCD RLS}\\ (general input)\end{tabular} &$M^2+2M+P_a$ &$\frac{3}{2}M^2+\frac{7}{2}M+1$ &0\\ \hline
        \begin{tabular}[c]{@{}l@{}} \text {(R) DCD RLS}\\ (tap-delayed input)\end{tabular}   &$3M+P_a$ &$5M+2$ &0\\
        \hline
    \end{tabular}
\end{table}

Table~\ref{table_4} mainly compares the complexity of robust (R) RLS-type with that of proposed (R) DCD variant in terms of ops, where we drop the calculation of $f_n$ dependent on a specific robust strategy. As in~\cite{zakharov2008RLS}, the DCD recursion requires $P_a=2N_uM + M_b$ additions at most for finding $\Delta \hat{\bm w}_n$. Thus, it is clear to see from Table~\ref{table_4}, for general input vector form, the DCD version reduces the complexity by at least a factor of 0.5 in contrast with the original algorithm, in terms of multiplications and additions. On the other hand, if the input vector $\bm x_n$ has a tapped-delay structure, i.e., $\bm x_n=[x_n,x_{n-1},...,x_{n-M+1}]^\text{T}$, where $x_n$ is a data sample at time~$n$, the calculation of $\bm R_n$ will be simplified. Specifically, assuming $f_n \approx f_{n-1}$, we can obtain the lower-right $(M-1)\times (M-1)$ block of $\bm R_n$ by copying the upper-left $(M-1)\times (M-1)$ block of $\bm R_{n-1}$. Then, considering the symmetry of $\bm R_n$, we only need the calculation of its first column:
\begin{equation}
\begin{array}{rcl}
\label{011}
[\bm R_n]_{:,1} \approx  \lambda [\bm R_{n-1}]_{:,1} + f_n x_n \bm x_n.
\end{array}
\end{equation}
Equation \eqref{011} is exact when $f_n=1$~\cite{zakharov2008RLS}. As claimed in~Section~II.~B, $f_n$ is normally close to 1, and becomes very small to suppress the update only when the impulsive noise happens. As such, using~\eqref{011} is also suitable for computing $\bm R_n$ in the proposed DCD recursion. In this scenario, the complexity is reduced to the same order of magnitude as that of LMS. This reduction is considerable especially for a long $\bm w^o$ such as in EC applications.

\subsection{Improving Tracking Performance}
For the proposed algorithms, there is also a trade-off between steady-state error and tracking capability for abrupt changes of $\bm w^o$, because of using the fixed forgetting factor~$\lambda$. To address this problem, one may utilize the adaptive combination (AC) of two independently running DCD-based filters. Like the AC-RL$p$N algorithm in~\cite{navia2012combination}, it combines RL$p$N filters with the large forgetting factor for low steady-state error and with the small one for good tracking capability. However, it requires at least double complexity of the original algorithm. Alternatively, the VFF has been also an effective mechanism for improving the original RLS algorithm~\cite{park1991fast,paleologu2008robust,cai2012low}. Consequently, to equip the proposed DCD-based algorithms, we also propose a simple VFF scheme:
\begin{equation}
\begin{array}{rcl}
\label{012}
\lambda_n = \lambda_\text{min} + (1-\lambda_\text{min}) \exp(-\rho e_{n,f}^2),
\end{array}
\end{equation}
where $\rho>0$ is a design parameter, $e_{n,f}^2$ is the impulse-free squared error which can be estimated by~\eqref{010}. As $n\rightarrow \infty$, $e_{n,f}^2$ converges to a small value, and according to \eqref{012}, $\lambda_n$ approaches 1, thus reducing the steady-state error. When $\bm w^o$ has a sudden change, $e_{n,f}^2$ becomes large due the mismatch estimation at that time, and $\lambda_n$ will approach a small forgetting factor $\lambda_\text{min}$, thus speeding up the convergence.

\section{Simulation results}
\begin{figure}[htb]
    \centering
    \includegraphics[scale=0.48] {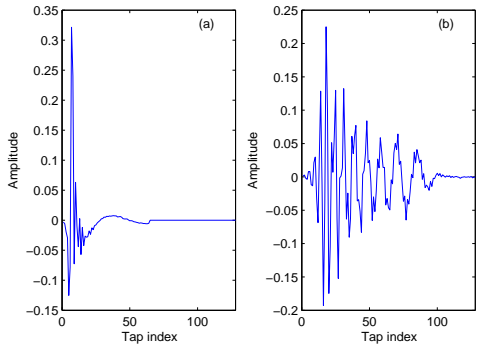}
    \hspace{2cm}\caption{Network echo channels: (a) sparse channel, (b) disperse channel.}
    \label{Fig1}
\end{figure}

\begin{figure}[htb]
    \centering
    \includegraphics[scale=0.5] {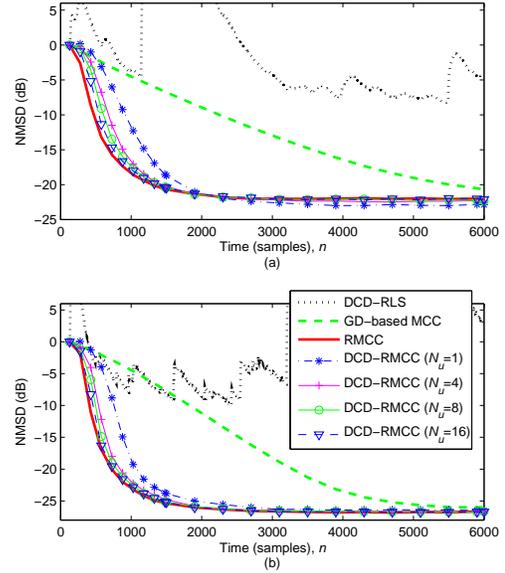}
    \hspace{2cm}\caption{NMSD curves of the DCD-RLS and MCC-based algorithms: (a)~sparse channel and correlated input; (b) disperse channel and white input. Parameters of algorithms are chosen as: $\lambda=0.998$ (all the algorithms); $\mu=0.001,\;\beta^2=0.6$ (GD-based MCC); $\beta^2=0.03$ (RMCC); $H=1,\; M_b=16$ (DCD).}
    \label{Fig2}
\end{figure}

In this section, simulations are conducted for identifying the network echo channel response $\bm w^o$ of length $M$ using an adaptive filter. The echo channels in Fig.~\ref{Fig1} are from the ITU-T G.168 standard, with $M=128$ taps~\cite{stnec2015}. For the tapped-delay input vector $\bm x_n$, its element $x_n$ is given by the first-order autoregressive model $x_n = \varrho x_{n-1} + \vartheta_n$, where $\vartheta_n$ is a zero-mean white Gaussian random process with unit variance. Both $\varrho=0$ (which is used only in Fig.~\ref{Fig2}(b)) and $\varrho=0.9$ correspond to the white and correlated inputs, respectively, with the eigenvalue spreads of 1 and 346. The $\alpha$-stable noise is set to $\alpha=1.4$ and $\gamma=1/20$. We use the normalized mean square deviation, $\text{NMSD}(n)= 10\log_{10}(||\bm w_n -\bm w^o||_2^2/||\bm w^o||_2^2)$, as a performance measure. All simulated curves are the average over 100 independent runs.

Fig.~\ref{Fig2} shows the NMSD performance of the DCD-RLS, GD-based MCC\footnote{The update equation is $\bm w_n = \bm w_{n-1}+\mu f_n e_n \bm x_n$ \cite{chen2014steady}.}, RMCC, and proposed DCD-RMCC algorithms. As expected for impulsive noise scenarios, the performance of the original DCD-RLS algorithm is poor, while the MCC-based algorithms are performing very well. The DCD-RMCC performance approaches that of the original RMCC algorithm as $N_u$ increases. In particular, $N_u=8\ll M$ (at most eight entries of $\bm w_n$ are updated per time $n$) has been enough for the DCD-RMCC performance to approach closely the RMCC performance regardless of whether $\bm w^o$ is sparse or not. However, as seen from Table~\ref{table_4}, the DCD-RMCC with $N_u=8$ reduces significantly the complexity of the RMCC. Although the DCD-RMCC requires 2.5 times multiplications of the GD-based MCC, the former (even if with $N_u=1$) has much faster convergence than the latter. Likewise, the convergence of the proposed low-cost DCD-RLM and DCD-RL$p$N versions also approximate well that of the RLM and RL$p$N algorithms, respectively; these results are omitted for brevity.

\begin{figure}[htb]
    \centering
    \includegraphics[scale=0.5] {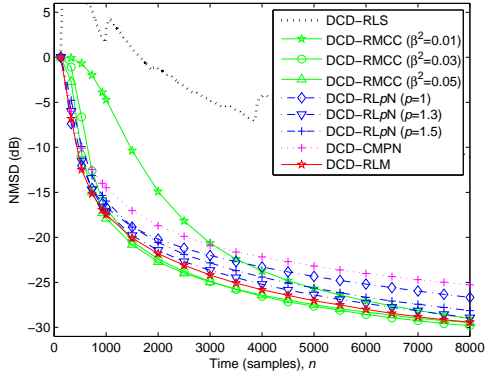}
    \hspace{2cm}\caption{NMSD curves of DCD-based RLS algorithms for the sparse channel. Parameters setting of algorithms is: $\lambda=0.9998$ (all the recursive algorithms); $\zeta=0.99,\;N_w=9$ (DCD-RLM). [$N_u=8$].}
    \label{Fig3}
\end{figure}

Fig.~\ref{Fig3} shows the NMSD of the proposed DCD-RMCC, DCD-RLM and DCD-RL$p$N algorithms, with $N_u=8$. The proposed algorithms show robustness in $\alpha$-stable noise and can arrive at similar performance by properly setting their parameters. This reason is they generally behave like the DCD-RLS and use a tiny $f_n$ to suppress the algorithms' adaptation once the impulsive noise appears. In addition, we also show the DCD-CMPN algorithm by applying the CMPN criteria in~\cite{zayyani2014continuous}, i.e., $\varphi(e)=\int_{1}^{2}|e|^pdp$ and $f(e)=\left( (2|e|-1)\ln(|e|)-|e|+1\right) /\left( |e|\ln^2(|e|)\right)$. For the $l_p$-norm based algorithms, $p$ should be slightly less than $\alpha$ in $\alpha$-stable noise; thus, the DCD-RL$p$N may outperform the DCD-CMPN, since the latter inherits the behavior of $p>\alpha$.

\begin{figure}[htb]
    \centering
    \includegraphics[scale=0.17] {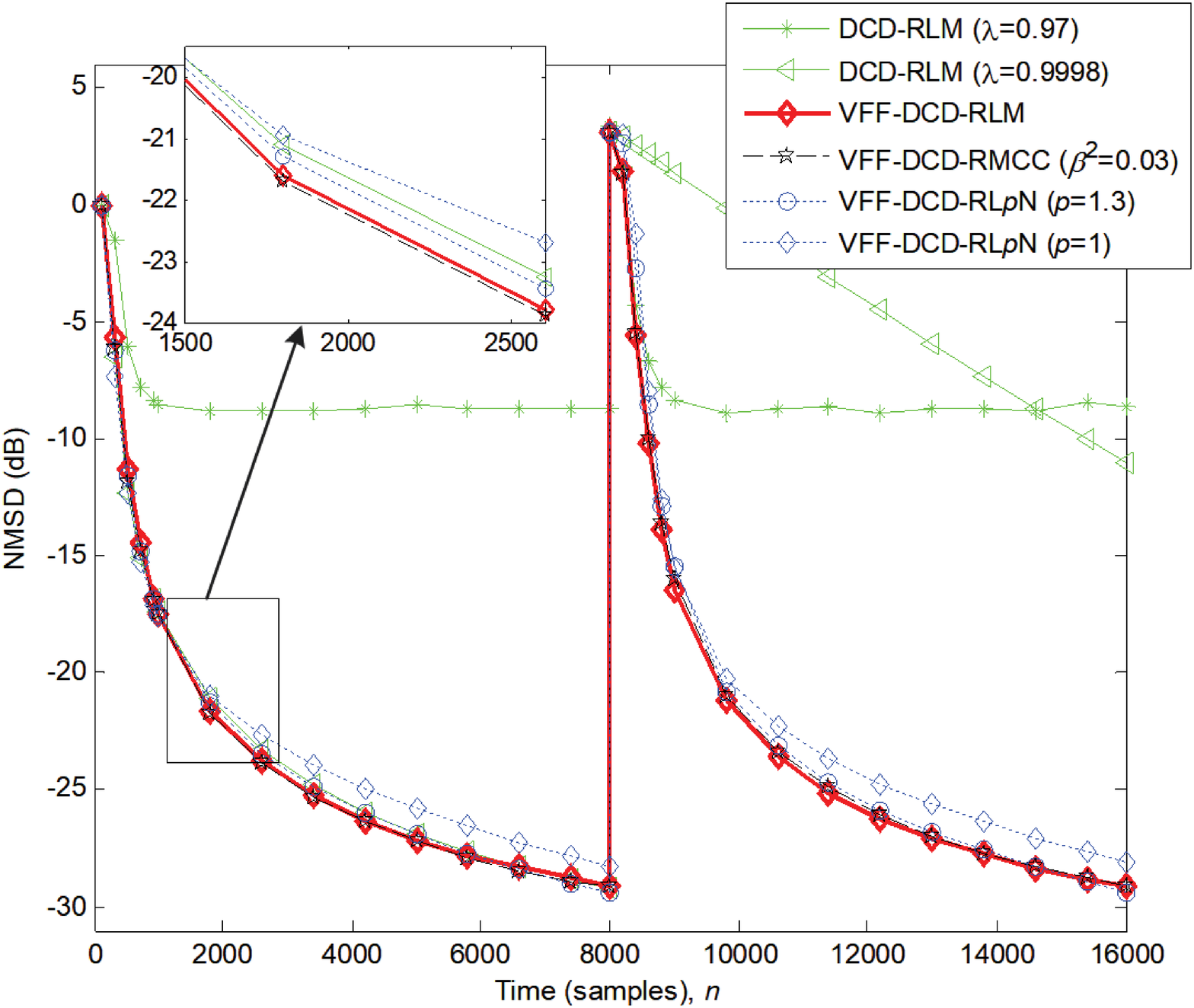}
    \hspace{2cm}\caption{NMSD curves of DCD-based RLS algorithms for the sparse channel. Parameters of VFF are: $\zeta=0.99,\;N_w=9,\;\rho=3,\;\lambda_\text{min}=0.97$.}
    \label{Fig4}
\end{figure}

Fig.~\ref{Fig4} demonstrates the tracking capability of the proposed algorithms, in a scenario where the echo channel changes at time $n=8001$ by shifting its impulse response by 12 samples. As one can see, using the proposed VFF instead of the fixed one, the DCD-based algorithms can reduce the steady-state error and improve the tracking capability.
\section{Conclusion}
We have proposed a general low-complexity recursion for developing RLS-type adaptive filtering algorithms operating in impulsive noise scenarios. This is based on using DCD iterations. As examples of the MCC, M-estimator, and $p$-norm strategies applied to this recursion, we have developed the DCD-RMCC, DCD-RLM, and DCD-RL$p$N algorithms, respectively. These algorithms show a performance similar to that of their high-complexity counterparts, RMCC, RLM, and RLpN algorithms, respectively. To improve the tracking capability of the algorithms, a simple time-varying forgetting factor mechanism has also been developed. Simulation results demonstrate the performance of the proposed algorithms.


\ifCLASSOPTIONcaptionsoff
  \newpage
\fi

\bibliographystyle{IEEEtran}
\bibliography{IEEEabrv,mybibfile}





%








\end{document}